%% file: iclr2027_conference.tex
\documentclass{article} 
\usepackage{iclr2027_conference,times}

\input{math_commands.tex}

\usepackage{hyperref}
\usepackage{url}
\usepackage{multirow}
\usepackage{graphicx}
\usepackage{makecell}
\usepackage{wrapfig}
\usepackage{needspace}
\usepackage{seqsplit}

\usepackage{booktabs}

\title{MiniWAM: Learning Compact Future Targets for Efficient World-Action Modeling}

\author{
Jie Chen$^{1,3}$,
Ruofei Bai$^{2,3}$,
Yuxin Cai$^{2,3}$,
Yifeng Zhang$^{1}$,
Chengyang He$^{1}$,
Jun Li$^{3}$, \\
\textbf{Wei-Yun Yau$^{3}$,
Guillaume Sartoretti$^{1}$}\\[4pt]
$^{1}$Department of Mechanical Engineering, National University of Singapore, Singapore\\
$^{2}$Nanyang Technological University, Singapore\\
$^{3}$A*STAR Institute of Advanced Intelligence and Computing (A*STAR IAIC), Singapore
}
\iclrfinalcopy 

\begin{document}
\maketitle
\lhead{}

\begin{abstract}

World modeling has emerged as an effective co-training objective for robot policies, giving rise to World Action Models (WAMs) that jointly predict actions and future states. 
However, most WAMs predict future states in the native representation space of pretrained visual backbones, resulting in high-dimensional targets with substantial training cost. 
We introduce MiniWAM, which instead predicts compact future representations learned from privileged current--future transitions. 
To construct these targets, we propose Predictive Representations via Inverse Spatiotemporal Modeling (PRISM), which combines inverse-dynamics supervision with feature reconstruction to emphasize control-relevant transition information while preserving useful future-state information. With the learned PRISM encoder frozen, MiniWAM is trained to jointly predict the resulting targets and robot actions from current observations.
With $65\times$ fewer native future feature tokens, MiniWAM consistently outperforms native future-feature prediction with both DINOv3 and WAN2.1 VAE features, while achieving up to an $8\times$ speedup in world--action training.
At 0.25B parameters, MiniWAM is already competitive with substantially larger WAMs on LIBERO, LIBERO-Plus, and RoboTwin 2.0 simulation benchmarks. Representation analyses further show that PRISM contributes behavioral structure beyond feature reconstruction alone. These results demonstrate that effective world--action modeling does not require predicting native visual futures, and that compact predictive representations provide a strong and substantially more efficient target for policy learning. The project page is available at: \href{https://j1dan.github.io/MiniWAM}{https://j1dan.github.io/MiniWAM}.

\end{abstract}

\section{Introduction}

Learning to predict future observations alongside robot actions provides a powerful training signal for robot control.
Recent World Action Models (WAMs) exploit this idea by jointly learning robot actions and future visual states, encouraging policy representations to capture environment dynamics relevant to control \citep{kim2026cosmos, ye2026world, yuan2026fast, yuan2026dreamwam, cen2025worldvla}. 
This paradigm has led to strong results across simulated and real-world manipulation. 
However, it also introduces a substantial computational burden. Most WAMs model the future through high-dimensional visual representations, typically dense video-VAE latents inherited from video-generation models \citep{wan2025wan}. 
Predicting these representations requires processing large spatiotemporal token sequences, making world modeling considerably more expensive. Yet preserving every detail in the native feature space may not be necessary for action learning.

Prior work has increasingly sought to improve the efficiency of WAMs by reducing the cost of future generation \citep{yuan2026fast}, sparsifying world--action interaction \citep{zhao2026faster}, or shrinking the model and token budget devoted to visual prediction \citep{li2026efficient}.
A complementary line of work replaces dense visual prediction with compact latent future supervision, using representations of future observations to shape policy-internal latent states \citep{zheng2025flare,su2026world,luo2026being}.
Together, these approaches suggest that the benefits of future modeling need not rely on explicit high-dimensional visual rollouts. 
However, existing efficiency-oriented WAMs still retain rich visual representations as their world-modeling targets, while latent approaches primarily incorporate future information through internal policy states or auxiliary conditioning spaces. 
This leaves a distinct question: can we directly learn a compact yet control-relevant future representation that replaces dense visual prediction as the world-modeling target of a WAM for efficient training?

\begin{figure*}[t]
    \centering
    \includegraphics[width=\textwidth]{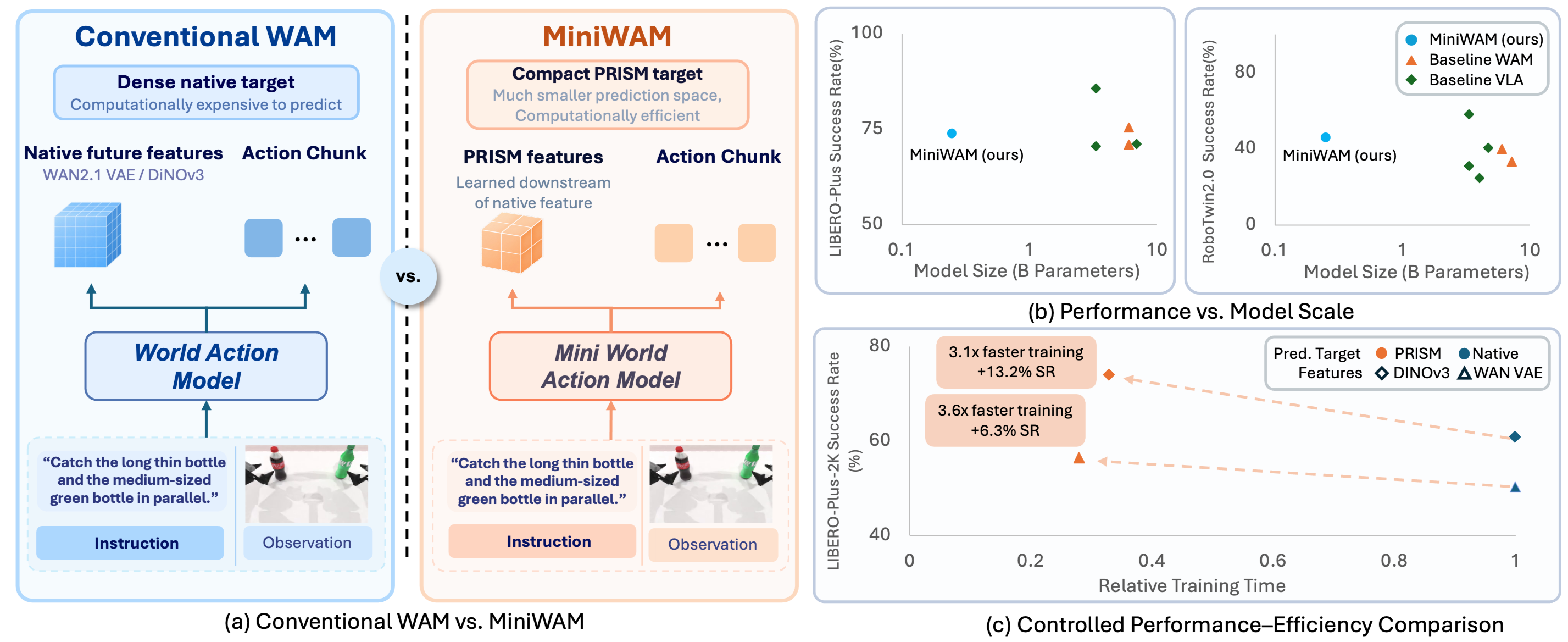}
    \caption{
    \textbf{MiniWAM replaces costly native future prediction with compact predictive representations.}
    (a) While conventional WAMs predict high-dimensional native future features, MiniWAM jointly predicts actions and compact PRISM targets learned from the same visual features via inverse dynamics and reconstruction.
    (b) MiniWAM remains competitive with substantially larger WAMs and VLAs on \textsc{LIBERO-Plus}.
    (c) In controlled comparisons across VAE and DINOv3 feature spaces, PRISM simultaneously improves success rate and reduces training time relative to native future-feature prediction.
    }
    \label{fig:headline}
\end{figure*}

To learn such a representation, we propose Predictive Representations via Inverse Spatiotemporal Modeling (PRISM). 
We aim to learn a compact representation that preserves information about both the actions underlying an observed transition and the resulting visual state.
During representation learning, a bottleneck encoder compresses current and future visual features into a compact set of latent tokens. 
Inverse-dynamics supervision encourages these tokens to retain transition information informative of the underlying actions, while feature reconstruction encourages them to preserve future-state content, without requiring the full detail of the native visual representation.

We then introduce \textbf{MiniWAM}, which separates learning \emph{what to represent} about the future from learning \emph{how to predict} it.
First, PRISM learns to encode current--future visual transitions into compact targets through inverse-dynamics supervision and feature reconstruction. 
The representation encoder is then frozen, and MiniWAM is trained to jointly predict the resulting PRISM target and robot actions given the current observation. 
In this way, MiniWAM changes what the world model branch is trained to predict, rather than merely reducing the cost of predicting an otherwise unchanged visual future.

PRISM is not tied to a particular visual feature space, because the bottleneck is learned downstream of the visual representation. We instantiate the framework with both generative WAN2.1 VAE features \citep{wan2025wan} and discriminative DINOv3 features \citep{simeoni2025dinov3}. 
By replacing dense future representations with a compact set of latent tokens, PRISM reduces the number of future tokens by $65\times$. 
Despite this substantial compression, MiniWAM consistently outperforms matched variants that directly predict the corresponding native future representations, while requiring substantially less training computation. 
These controlled comparisons show that the benefit does not arise simply from choosing a stronger visual encoder: replacing its dense future representation with the compact PRISM target improves the performance--efficiency trade-off.
This performance--efficiency advantage is illustrated in Figure~\ref{fig:headline}.

Empirically, MiniWAM reduces WAM training cost by up to 8$\times$ while outperforming native future-feature prediction across simulation benchmarks. 
A 0.25B MiniWAM is already competitive with WAMs built upon substantially larger pretrained video-generation backbones. Ablations show that adding inverse-dynamics supervision to reconstruction-only bottlenecks consistently improves performance across bottleneck capacities.
These results suggest that strong world--action modeling does not require predicting the native visual future: compact learned representations can provide a stronger and substantially more efficient target for predictive co-training.

\section{Related Work}

\paragraph{World Models and World--Action Models for Robot Control.}
World models learn predictive representations of environment dynamics to support planning and policy learning, often in latent or pretrained feature spaces rather than pixel space \citep{NEURIPS2018_2de5d166, Hafner2020Dream,zhou2025dinowm}. 
Building on this principle, recent methods increasingly integrate future prediction directly into policy training, giving rise to World Action Models (WAMs) that jointly model future states and robot actions \citep{kim2026cosmos,ye2026world,yuan2026fast}.
Using future prediction as a co-training objective, WAMs encourage the policy to capture structured environment dynamics while sharing information between world and action streams. This formulation has achieved strong manipulation performance, but typically requires predicting dense spatiotemporal visual representations, such as video-VAE latents, substantially increasing the cost of policy training.

\paragraph{Efficient and Compact Future Modeling.}
Recent work improves WAM efficiency through architectural simplification and reduced future-modeling computation \citep{yuan2026fast,zhao2026faster,li2026efficient}. 
A complementary line replaces dense future prediction with compact latent supervision, aligning or supervising policy-internal representations with features derived from future observations \citep{zheng2025flare,luo2026being,su2026world}. 
These approaches show that predictive information can be represented more compactly than the original visual future, but typically retain native visual features as world-modeling targets or use compact representations primarily as internal policy states or conditioning signals. 
Concurrent work \citep{wang2026openwam} shows that channel-wise compression of semantic prediction targets can yield strong world--action co-training performance. 
PRISM instead learns a control-relevant transition representation as the world-prediction target, jointly compressing its spatial and temporal dimensions to substantially reduce the number of predicted tokens and the cost of world–action co-training.

\section{Mini World Action Model with Predictive Representations}
\label{sec:method}

\subsection{Overview}

Given a current observation $o_t$, a future observation window $o_{t+1:t+H_\mathrm{v}}$, and the corresponding action chunk $\mathbf{a}_t=(a_t,\ldots,a_{t+H_\mathrm{a}-1})$, we seek a compact representation of the visual transition that can replace native future features as the predictive target of a World Action Model. Let $f$ denote a frozen visual encoder, and let $(x_t,X_t^+)$ denote the current and future feature blocks obtained by encoding $o_{t:t+H_{\mathrm v}}$ with $f$. PRISM introduces a spatiotemporal encoder $E_\phi$ that maps these privileged current--future features to a compact predictive representation $L_t=E_\phi(x_t,X_t^+)$, whose dimensionality is substantially smaller than that of the native future representation $X_t^+$.

\begin{figure*}[t]
    \centering
    \includegraphics[width=\textwidth]{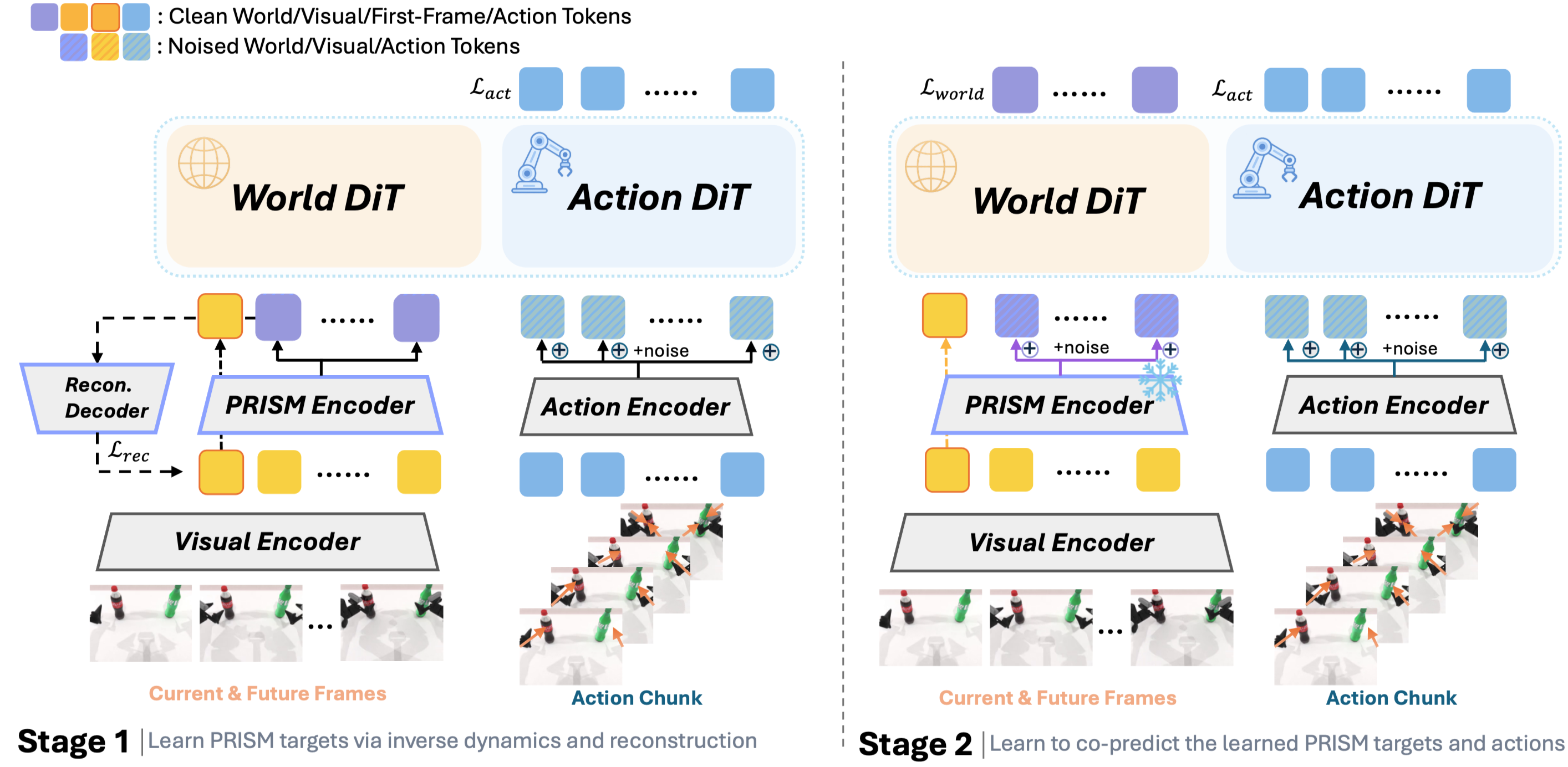}
    \vspace{-15pt}
    \caption{
    \textbf{Two-stage training pipeline of PRISM and MiniWAM.}
    In \textbf{Stage 1}, privileged current--future visual features are
    compressed by the PRISM encoder into a compact predictive representation
    $L_t$, shaped through inverse action prediction and feature reconstruction.
    The corresponding world tokens remain clean while the action tokens are
    corrupted for conditional action prediction.
    In \textbf{Stage 2}, the PRISM encoder is frozen and provides compact
    world-modeling targets; MiniWAM jointly denoises the world and action
    tokens using its world and action experts.
    Solid and hatched tokens denote clean and noised tokens, respectively.
    Language and proprioceptive conditioning are omitted for clarity.
    }
    \label{fig:miniwam}
\end{figure*}

Training comprises two stages, as illustrated in Figure~\ref{fig:miniwam}. In \emph{Stage 1}, PRISM learns $E_\phi$ from future transitions using inverse-dynamics supervision and feature reconstruction: the former biases $L_t$ toward control-relevant dynamics, while the latter preserves useful future-state information. In \emph{Stage 2}, $E_\phi$ is frozen and $L_t$ becomes the world-modeling target of MiniWAM, which jointly predicts the PRISM representation and robot actions from the current observation and conditioning context. Future observations are required only to construct training targets and are unavailable at deployment.

Both stages use a Mixture-of-Transformers (MoT) composed of two modality-specific diffusion transformer (DiT) \citep{peebles2023scalable} experts: a world expert that models the PRISM representation $L_t$ and an action expert that models $\mathbf{a}_t$. We refer to their tokenized inputs as \emph{world tokens} and \emph{action tokens}, respectively. Each expert retains its own input/output projections, feed-forward layers, and timestep modulation, while mixed self-attention enables information exchange between the two modalities. In Stage~1, the world tokens remain clean and condition action prediction; in Stage~2, the world and action tokens are jointly denoised.

We use flow matching \citep{lipman2023flow} for both prediction branches. For
$y\in\{L,\mathbf{a}\}$, we sample Gaussian noise $\epsilon_y$ and noise level
$\tau_y$ and construct

$$
\widetilde y_{\tau_y}
=
(1-\sigma_{\tau_y})y+\sigma_{\tau_y}\epsilon_y,
\qquad
u_y^\star=\epsilon_y-y.
$$

The unified objective is
$\mathcal{L}_{\mathrm{FM}}
=
\lambda_L \mathbb{E}\!\left[w(\tau_L)\|\hat u_L-u_L^\star\|_2^2\right]
+
\lambda_a \mathbb{E}\!\left[w(\tau_a)\|\hat u_a-u_a^\star\|_2^2\right]$.

Stage 1 keeps $L_t$ clean and sets $\lambda_L=0$, using it for inverse action prediction. Stage 2 independently corrupts both $L_t$ and $\mathbf{a}_t$ and optimizes both prediction terms.

\subsection{Stage 1: Learning the Predictive Representation}

\paragraph{Spatiotemporal transition bottleneck.} 

The transition encoder $E_\phi$ is instantiated as a lightweight 3D convolutional network. It jointly processes the current and future visual features $x_t$ and $X_t^+$ arranged along the temporal dimension, progressively compressing them via strided 3D convolutions into a compact spatiotemporal grid $L_t\in\mathbb{R}^{T_L\times H_L\times W_L\times D_L}$, which is subsequently flattened into world tokens for the MoT.
We parameterize the bottleneck as a diagonal Gaussian,

\[
q_\phi(L_t\mid x_t,X_t^+)
=
\mathcal{N}\!\left(
\mu_\phi(x_t,X_t^+),
\operatorname{diag}(\sigma_\phi^2(x_t,X_t^+))
\right),
\qquad
L_t=\mu_\phi+\sigma_\phi\odot\epsilon,
\quad
\epsilon\sim\mathcal{N}(0,I).
\]

PRISM shapes this bottleneck using two complementary objectives. For inverse dynamics, $L_t$ is provided as a clean conditioning sequence while the action sequence is corrupted according to the flow-matching process above. Direct current-observation and language conditioning are removed from the MoT, forcing visual information relevant to action prediction to pass through $L_t$. The current feature $x_t$ remains an input to $E_\phi$, allowing the bottleneck to represent the current--future transition rather than the future state in isolation.

Feature reconstruction encourages the compact representation to preserve useful information about the native future state. A lightweight decoder $D_\psi$ reconstructs $X_t^+$ using $\mathcal{L}_{\mathrm{rec}}=\|D_\psi(L_t)-X_t^+\|_2^2$, while a weak KL term regularizes the bottleneck distribution. The Stage-1 objective is

$$
\mathcal{L}_{\mathrm{S1}}
=
\lambda_a
\mathbb{E}
\left[
w(\tau_a)
\|\hat u_a-u_a^\star\|_2^2
\right]
+
\lambda_{\mathrm{rec}}\mathcal{L}_{\mathrm{rec}}
+
\beta
D_{\mathrm{KL}}
\left[
q_\phi(L_t\mid x_t,X_t^+)
\,\|\,\mathcal{N}(0,I)
\right].
$$

Inverse dynamics therefore biases the compact state toward behaviorally meaningful transition information, while reconstruction preserves predictive information about the future visual state. The decoder is used only during Stage 1 and is discarded afterward.

\subsection{Stage 2: World--Action Modeling with PRISM Targets}

After Stage~1, we freeze $E_\phi$ and use its posterior mean
$L_t^\star=\mu_\phi(x_t,X_t^+)$ as a deterministic world-modeling target.
MiniWAM receives no future observations as input and learns to predict
$L_t^\star$ from the current observation while jointly predicting the action
trajectory. The MoT sequence is
$[x_t^0,\widetilde L_{\tau_L},\widetilde{\mathbf a}_{\tau_a}]$, where
$x_t^0$ denotes clean current-observation tokens, and
$\widetilde L_{\tau_L}$ and $\widetilde{\mathbf a}_{\tau_a}$ denote the
noised world and action tokens at noise levels $\tau_L$ and $\tau_a$,
respectively. Both branches use independent Gaussian noise and, by default,
independently sampled noise levels. Language instructions and proprioceptive states are provided as contextual conditioning via cross-attention.

Stage 2 also changes the interaction between the two prediction branches. The world and action tokens attend to one another bidirectionally, allowing world and action predictions to be refined jointly. Both terms of the unified objective are now active:

$$
\mathcal{L}_{\mathrm{S2}}
=
\lambda_L
\mathbb{E}
\left[
w(\tau_L)
\|\hat u_L-u_L^\star\|_2^2
\right]
+
\lambda_a
\mathbb{E}
\left[
w(\tau_a)
\|\hat u_a-u_a^\star\|_2^2
\right].
$$

Thus, MiniWAM retains the standard world--action co-training formulation while replacing native future features with the compact PRISM target $L_t^\star$.

\paragraph{Inference.} At deployment, $E_\phi$ and future observations are no longer needed. Given only the current observation, instruction, and proprioception, we initialize $L^{(K)}\sim\mathcal{N}(0,I)$ and $\mathbf{a}^{(K)}\sim\mathcal{N}(0,I)$ and jointly denoise them through the world and action experts. Each MoT forward pass predicts both velocity fields, after which the two variables are updated according to their respective flow schedules. The final action chunk is executed in a receding-horizon manner.

\subsection{Visual Representation Instantiations}

PRISM operates downstream of the visual encoder and therefore does not assume a particular representation type. We study two distinct feature families. \textbf{VAE features.} We use a frozen WAN2.1 video VAE, representative of the generative latent spaces commonly modeled by video-based WAMs. Its current and future features form a spatiotemporal volume that is compressed by $E_\phi$. \textbf{DINOv3 features.} We use frozen DINOv3 patch features, temporally aggregating the framewise spatial features before $E_\phi$ compresses the complete current--future sequence into the same compact spatiotemporal representation.

These two instantiations keep the PRISM formulation unchanged while varying the underlying visual feature space. This allows us to test whether learning a compact, control-relevant predictive target is beneficial across both generative and discriminative visual representations.

\section{Experiments}
\label{sec:experiments}

We evaluate PRISM to answer three key questions:
\textbf{(Q1)} Can PRISM replace native future-feature prediction while improving the performance--training-efficiency trade-off?
\textbf{(Q2)} How does MiniWAM compare with state-of-the-art VLAs and WAMs? 
\textbf{(Q3)} What aspects of future dynamics does PRISM encourage its representation to capture?

\subsection{Experimental Setup}

We evaluate policy success on \textsc{LIBERO} \citep{liu2023libero}, \textsc{LIBERO-Plus} \citep{Fei_2026_CVPR}, and \textsc{RoboTwin 2.0} \citep{chen2026robotwin}, covering controlled world-modeling comparisons, robustness under distribution shifts, and multi-task bimanual manipulation. We additionally benchmark training efficiency on LIBERO and RoboTwin 2.0. Detailed training configurations, evaluation protocols, and hyperparameters are provided in the Appendix.

\paragraph{Benchmarks.} 
\textbf{\textsc{LIBERO}.} \textsc{LIBERO} comprises four standard suites---\textsc{LIBERO-Spatial}, \textsc{LIBERO-Object}, \textsc{LIBERO-Goal}, and \textsc{LIBERO-Long}---with 40 language-conditioned manipulation tasks, and serves as our primary benchmark for controlled comparisons of world-modeling targets. \textbf{\textsc{LIBERO-Plus}.} \textsc{LIBERO-Plus} extends these tasks with systematic perturbations to camera viewpoints, robot initial states, language instructions, lighting, backgrounds, observation noise, and object layouts, allowing us to evaluate robustness under controlled distribution shifts. \textbf{\textsc{RoboTwin}~2.0.} \textsc{RoboTwin}~2.0 is a 50-task bimanual manipulation benchmark with the Aloha-AgileX robot; we use its multi-task setting and evaluate policy success under both clean and randomized conditions. We additionally use \textsc{RoboTwin}~2.0 for training-efficiency measurements, where its additional wrist view increases the visual prediction workload relative to \textsc{LIBERO}.

\subsection{PRISM versus Direct Future-Feature Prediction}
\label{sec:exp_feature}

We first study whether the learned PRISM representation can replace native future features as the co-training target of a WAM. We construct controlled variants for both visual representation families considered in this work. For \textbf{Native Co-Train}, the world expert directly predicts the native future representation produced by the corresponding visual encoder: future VAE latent blocks for the VAE variant and the complete sequence of future DINOv3 patch features for the DINOv3 variant. For \textbf{PRISM}, the native future features are further compressed by the learned transition encoder, and the resulting compact representation serves as the world-modeling target. We additionally include a \textbf{w/o Co-Train} variant for each visual feature family, where the world-prediction objective is disabled. This baseline isolates the contribution of predictive co-training from that of the underlying visual representation. For a controlled comparison, the world expert is adapted only to the dimensionality and tokenization of each target, while the training data, Stage-2 optimization schedule, and number of policy-training updates are kept fixed. PRISM's Stage 1 learns the representation used as its prediction target. All variants are trained on \textsc{LIBERO} and evaluated under matched protocols on \textsc{LIBERO} and \textsc{LIBERO-Plus-2K}.

\begin{table}[t]
\centering
\caption{\textbf{Controlled comparison of world-modeling targets on
\textsc{LIBERO} and \textsc{LIBERO-Plus-2K}.}
All variants use the same 0.25B world--action architecture and training
protocol within each visual feature family.}
\label{tab:feature_comparison}
\begin{tabular}{@{}clcc@{}}
\toprule
Visual features
& Co-training target
& \textsc{LIBERO}
& \textsc{LIBERO-Plus-2K} \\
&
&
SR (\%) $\uparrow$
& SR (\%) $\uparrow$ \\
\midrule

\multirow[c]{3}{*}{VAE}
& w/o Co-Train
& 62.3
& 25.1 \\

& w/ Native Co-Train
& 92.8
& 50.2 \\

& \textbf{w/ PRISM}
& \textbf{94.8}
& \textbf{56.5} \\

\midrule

\multirow[c]{3}{*}{DINOv3}
& w/o Co-Train
& 89.0
& 56.5 \\

& w/ Native Co-Train
& 93.2
& 60.9 \\

& \textbf{w/ PRISM}
& \textbf{97.2}
& \textbf{74.1} \\

\bottomrule
\end{tabular}
\vspace{-20pt}
\end{table}

Table~\ref{tab:feature_comparison} yields two observations. First, native future-feature co-training substantially improves control across both visual representation families: it raises the VAE-based policy from 62.3\% to 92.8\% and the DINOv3-based policy from 89.0\% to 93.2\%. Second, replacing native future features with the learned PRISM representation further improves performance, reaching 94.8\% with VAE features and 97.2\% with DINOv3 features. Under matched Stage-2 training, these gains demonstrate the benefit of the learned PRISM target. On \textsc{LIBERO-Plus-2K}, PRISM also improves over native co-training by 6.3 and 13.2 percentage points for VAE and DINOv3, respectively. Notably, PRISM improves over native co-training for both generative VAE features and discriminative DINOv3 features, demonstrating that its benefit is not tied to a particular form of visual representation.

\paragraph{Training efficiency.}
The performance gain above is accompanied by a substantial reduction in world-modeling computation, as reported in Table~\ref{tab:training_efficiency}. Direct prediction requires the world expert to operate on dense native future representations, whereas PRISM compresses the current--future transition into a compact spatiotemporal target, reducing the world-token burden. We benchmark matched PRISM and direct-prediction variants at future horizons of 16 and 32 frames on both \textsc{LIBERO} and \textsc{RoboTwin}~2.0 using the same global batch size and hardware. \textsc{RoboTwin}~2.0 additionally includes a second wrist camera view, providing a heavier multi-view prediction workload.

PRISM consistently reduces predictive co-training cost across both benchmarks and visual feature spaces, with larger gains under heavier prediction workloads. At horizon 32, estimated total training speedup reaches $8.46\times$ on \textsc{RoboTwin}~2.0, while Stage-2 training alone reaches $10.48\times$. Together with Table~\ref{tab:feature_comparison}, these results show that learned PRISM targets improve the performance--efficiency trade-off of world--action co-training.

\begin{table}[t]
\centering
\caption{\textbf{Training efficiency of PRISM versus native co-training.}
Throughput is measured in samples/s with global batch 256 under matched
hardware. S2 is Stage-2 speedup; Total estimates training speedup including Stage 1, assuming equal per-update costs for the two PRISM stages (Appendix~\ref{app:efficiency_accounting}).}
\label{tab:training_efficiency}
\small
\begin{tabular}{@{}llcccccc@{}}
\toprule
& &
\multicolumn{3}{c}{\textbf{Horizon 16}} &
\multicolumn{3}{c}{\textbf{Horizon 32}} \\
\cmidrule(lr){3-5}
\cmidrule(lr){6-8}

\textbf{Visual Feature} &
\textbf{Target} &
\textbf{Throughput} &
\textbf{S2} &
\textbf{Total} &
\textbf{Throughput} &
\textbf{S2} &
\textbf{Total} \\
& &
$\uparrow$ &
$\uparrow$ &
$\uparrow$ &
$\uparrow$ &
$\uparrow$ &
$\uparrow$ \\
\midrule

\multicolumn{8}{c}{\textit{\textsc{LIBERO}}} \\
\midrule

WAN2.1 VAE
& Native
& 30.81
& 1.00$\times$
& 1.00$\times$
& 14.20
& 1.00$\times$
& 1.00$\times$ \\

& \textbf{PRISM}
& \textbf{142.04}
& \textbf{4.61$\times$}
& \textbf{3.55$\times$}
& \textbf{124.92}
& \textbf{8.80$\times$}
& \textbf{6.77$\times$} \\

\addlinespace
DINOv3
& Native
& 36.17
& 1.00$\times$
& 1.00$\times$
& 15.55
& 1.00$\times$
& 1.00$\times$ \\

& \textbf{PRISM}
& \textbf{143.25}
& \textbf{3.96$\times$}
& \textbf{3.05$\times$}
& \textbf{126.09}
& \textbf{8.11$\times$}
& \textbf{6.24$\times$} \\

\midrule

\multicolumn{8}{c}{\textit{\textsc{RoboTwin} 2.0}} \\
\midrule

WAN2.1 VAE
& Native
& 17.70
& 1.00$\times$
& 1.00$\times$
& 7.79
& 1.00$\times$
& 1.00$\times$ \\

& \textbf{PRISM}
& \textbf{93.77}
& \textbf{5.30$\times$}
& \textbf{4.27$\times$}
& \textbf{81.66}
& \textbf{10.48$\times$}
& \textbf{8.46$\times$} \\

\addlinespace
DINOv3
& Native
& 21.48
& 1.00$\times$
& 1.00$\times$
& 8.61
& 1.00$\times$
& 1.00$\times$ \\

& \textbf{PRISM}
& \textbf{93.57}
& \textbf{4.36$\times$}
& \textbf{3.51$\times$}
& \textbf{81.47}
& \textbf{9.46$\times$}
& \textbf{7.63$\times$} \\

\bottomrule
\end{tabular}
\vspace{-10pt}
\end{table}

\vspace{-5pt}

\subsection{Comparison with State-of-the-Art Robot Policies} \label{sec:exp_sota} 

We compare MiniWAM with state-of-the-art VLAs \citep{OpenVLA, PI0, FAST, PI05, UniVLA, community2026starvla,cai2026xiaomi} and WAMs \citep{cen2025worldvla, yuan2026fast, yuan2026dreamwam, cai2026aha}. 
The results are summarized in Table~\ref{tab:sota}.
For unmarked rows, we evaluate released checkpoints on \textsc{LIBERO} and \textsc{LIBERO-Plus} using the same task configurations and random seeds as MiniWAM. Rows marked with $\dagger$ use externally reported results. On \textsc{RoboTwin}~2.0, we report the official clean2clean/clean2random leaderboard mean. Appendix~\ref{app:sota_provenance} records the source and evaluation setting of every externally reported entry.

\begin{table}[t]
\centering
\caption{\textbf{Comparison with state-of-the-art robot policies.}
$\dagger$ denotes externally reported results. Scale is the reported model size;
Embodied PT and Backbone Init. denote embodied pretraining and backbone
initialization.
``--'' means no pretrained backbone and ``-'' means unavailable.}
\label{tab:sota}

\small
\setlength{\tabcolsep}{3.2pt}
\renewcommand{\arraystretch}{1.08}

\begin{tabular}{@{}llcccccc@{}}
\toprule
Type
& Method
& Scale
& \makecell{Embodied\\PT.}
& \makecell{Backbone\\Init.}
& \makecell{\textsc{LIBERO}\\SR (\%) $\uparrow$}
& \makecell{\textsc{LIBERO-Plus}\\SR (\%) $\uparrow$}
& \makecell{\textsc{RoboTwin 2.0}\\SR (\%) $\uparrow$} \\
\midrule

\multirow[c]{7}{*}{VLA}
& OpenVLA-OFT
& 7B
& $\checkmark$
& VLM
& 96.6
& 71.4
& - \\

& $\pi_0$
& 3.3B
& $\checkmark$
& VLM
& 93.5
& 70.5
& 31.4 \\

& $\pi_0$-FAST$^\dagger$
& 3.3B
& $\checkmark$
& VLM
& 85.5
& 61.6
& - \\

& UniVLA$^\dagger$
& 7B
& $\checkmark$
& VLM
& 95.2
& 42.9
& - \\

& StarVLA-OFT$^\dagger$
& 4B
& $\times$
& VLM
& 96.6
& -
& 24.8 \\

& Xiaomi-Robotics-0$^\dagger$
& 4.7B
& $\checkmark$
& VLM
& 98.7
& -
& 40.5 \\

& $\pi_{0.5}$
& 3.3B
& $\checkmark$
& VLM
& 96.4
& 85.7
& 58.4 \\

\midrule

\multirow[c]{5}{*}{WAM}
& WorldVLA$^\dagger$
& 7B
& $\times$
& VLM
& 79.1
& 25.0
& - \\

& Fast-WAM
& 6B
& $\times$
& VGM
& 95.8
& 71.0
& 39.9 \\

& DreamWAM
& 6B
& $\times$
& VGM
& 96.9
& 75.4
& - \\

& AHA-WAM$^\dagger$
& 7.2B
& $\times$
& VGM
& -
& -
& 33.8 \\

& MiniWAM (ours)
& 0.25B
& $\times$
& --
& 97.2
& 73.6
& 46.1 \\


\bottomrule
\end{tabular}
\vspace{-10pt}
\end{table}

Despite its substantially smaller scale, MiniWAM remains competitive with much larger WAMs on both benchmarks. On \textsc{LIBERO}, the 0.25B MiniWAM model reaches 97.2\% success, comparable to the 6B-scale Fast-WAM and DreamWAM models. More importantly, PRISM retains strong performance under the distribution shifts in \textsc{LIBERO-Plus}, achieving 73.6\% success and outperforming Fast-WAM while approaching DreamWAM, despite using roughly an order of magnitude fewer world--action model parameters and no pretrained video-generation backbone. On \textsc{RoboTwin}~2.0, MiniWAM reaches 46.1\%, exceeding
Fast-WAM and AHA-WAM. These results indicate that scaling the video model is not the only route to strong world--action learning: a compact model can remain highly competitive when its predictive objective is defined over an appropriately structured future representation.

\subsection{Ablation Studies}

We study how PRISM benefits from inverse-dynamics supervision and how its performance changes as the predictive bottleneck is compressed. All ablations use DINOv3 features and are evaluated on a fixed stratified
2K subset of \textsc{LIBERO-Plus}, covering suites, perturbation categories, and difficulty levels. Here, $n\times n$ denotes the spatial resolution of the learned bottleneck token grid, and $D$ denotes the feature dimension of each token.

\begin{wraptable}{r}{0.48\textwidth}
\vspace{-20pt}
\centering
\caption{\textbf{PRISM ablation on \textsc{LIBERO-Plus-2K}.}
Success rate (\%).}
\label{tab:prism_ablation}
\small
\setlength{\tabcolsep}{4pt}
\begin{tabular}{@{}lccc@{}}
\toprule
Bottleneck & Recon. & Inv. dyn. & Both \\
\midrule
$4{\times}4$, $D{=}64$ & 69.5 & 70.5 & \textbf{74.1} \\
$2{\times}2$, $D{=}64$ & 63.3 & 67.5 & \textbf{67.8} \\
$1{\times}1$, $D{=}16$ & 57.0 & 56.8 & \textbf{61.7} \\
\bottomrule
\end{tabular}
\vspace{-6pt}
\end{wraptable}
Table~\ref{tab:prism_ablation} reveals two consistent trends. First, neither inverse dynamics nor reconstruction alone uniformly dominates, whereas their combination performs best across all bottleneck capacities. Adding inverse-dynamics supervision to reconstruction improves success rate by roughly 4.6 percentage points across all capacities, indicating that it provides complementary control-relevant supervision beyond feature reconstruction. In Section~\ref{sec:bottleneck_analysis}, we further examine how this supervision shapes the learned representation geometry. Second, performance degrades as the bottleneck becomes restrictive, with PRISM success rate decreasing from 74.1\% at the largest bottleneck to 61.7\% at the smallest. Despite this compression, the combined PRISM objective consistently outperforms reconstruction alone.

\subsection{Spatiotemporal Bottleneck Analysis}
\label{sec:bottleneck_analysis}

We next investigate what the spatiotemporal encoder learns under PRISM supervision. In particular, we ask whether the bottleneck reorganizes heterogeneous visual features into a compact representation structured around behaviorally relevant transitions. \textbf{Setup.}
We conduct this analysis on \textsc{LIBERO-90}, whose 90 task IDs are disjoint from the \textsc{LIBERO-40} tasks used to train the bottleneck encoder. We construct a balanced set of 4,500 transitions spanning all 90 tasks and 10 episode-progress bins, with a fixed balanced subset of 1,800 transitions used for CKA analysis. For every comparison, the native Wan-VAE and DINOv3 representations and their corresponding PRISM bottlenecks are constructed from the same physical current-to-future observation window $x_{t:t+H}$. Thus, both sides have access to identical temporal observations. The native transition features are spatially averaged at each encoder-native timestep, while the primary PRISM representation retains the full $3\times4\times4$ posterior grid $L_\mu$. We analyze the main and wrist camera views separately.

\paragraph{Does PRISM align transition-specific geometry?} Wan-VAE and DINOv3 encode the same visual transition with substantially different inductive biases: the former uses a generative latent space, whereas the latter provides discriminative visual features. We ask whether PRISM transforms these heterogeneous representations into a more consistent relational geometry across backbones, particularly for variation associated with the underlying transition rather than task or scene identity.

\begin{wraptable}{r}{0.49\textwidth}
\vspace{-20pt}
\centering
\caption{\textbf{Cross-backbone alignment on unseen
\textsc{LIBERO-90} tasks.}
Entries report main/wrist-view CKA.}
\label{tab:bottleneck_cka}
\small
\setlength{\tabcolsep}{4pt}
\begin{tabular}{@{}lccc@{}}
\toprule
Protocol & Raw & Random & PRISM \\
\midrule
Global
& .888/.571 & .891/.469 & .889/\textbf{.736} \\
Within-task
& .372/.249 & .682/.347 & \textbf{.795}/\textbf{.705} \\
Deconfounded
& .109/.071 & .162/.162 & \textbf{.379}/\textbf{.393} \\
\bottomrule
\end{tabular}
\vspace{-6pt}
\end{wraptable}

We measure cross-backbone similarity using debiased linear centered kernel alignment (CKA) \citep{cka} on exactly paired transitions. Because global CKA may be dominated by task- and scene-level structure, we additionally consider two conditioned variants that reduce shared task and current-observation structure. \emph{Global} CKA measures overall representational similarity, while \emph{within-task} CKA subtracts each task's mean feature vector separately for each representation, then computes a single CKA score over all task-centered transitions. Finally, \emph{task/current-deconfounded} CKA removes variation that is linearly predictable from task identity, episode progress, and the current observation before measuring the remaining cross-backbone alignment. All nuisance models are fit using cross-fitting at the demonstration level to prevent information leakage from evaluated trajectories. Further implementation details are provided in Appendix~\ref{app:analysis_cka}.

Table~\ref{tab:bottleneck_cka} shows that PRISM substantially strengthens cross-backbone alignment once task and current-state confounds are controlled. While global CKA provides little separation on the main view, PRISM markedly improves within-task and deconfounded alignment over both raw transitions and the architecture-matched random encoder. The effect is particularly pronounced for wrist observations, where within-task CKA increases from $0.249$ to $0.705$ and deconfounded CKA from $0.071$ to $0.393$. These results suggest that PRISM learns a shared transition geometry that is more consistent across fundamentally different visual backbones.

\paragraph{Does the learned geometry reflect behavior?} Greater transition-specific agreement alone does not establish that the resulting geometry is meaningful for control. We therefore examine the local neighborhoods induced by each representation using a locked split of 882 episode-disjoint queries and 3,618 gallery transitions. For every query, we retrieve its nearest neighbors while excluding all samples from the same task. This leave-task-out protocol prevents task identity from trivially determining the retrieved examples and tests whether similar representations correspond to similar transitions across different manipulation tasks. 

We first measure whether Wan- and DINO-based representations retrieve the same transitions using Jaccard similarity over their top-$5$ neighbor sets. Behavioral metrics are computed separately over the Wan- and DINO-retrieved top-$5$ neighbors and then symmetrically averaged across the two retrieval directions. Action RMSE is measured over normalized 16-step action chunks; EEF error is the Euclidean difference between current-to-future end-effector displacement vectors in metres; and gripper agreement is the fraction of action timesteps with matching gripper-command signs.

\begin{wraptable}{r}{0.49\textwidth}
\vspace{-20pt}
\centering
\caption{\textbf{Behavioral structure on unseen
\textsc{LIBERO-90} tasks.}
Leave-task-out retrieval results are reported as Raw $\rightarrow$ PRISM.}
\label{tab:bottleneck_retrieval}
\small
\setlength{\tabcolsep}{4pt}
\begin{tabular}{@{}lcc@{}}
\toprule
Metric & Main & Wrist \\
\midrule
Jaccard@5 $\uparrow$
& 0.170 $\rightarrow$ \textbf{0.512}
& 0.107 $\rightarrow$ \textbf{0.333} \\

Action RMSE $\downarrow$
& 0.400 $\rightarrow$ \textbf{0.365}
& 0.370 $\rightarrow$ \textbf{0.272} \\

EEF error (m) $\downarrow$
& 0.083 $\rightarrow$ \textbf{0.072}
& 0.089 $\rightarrow$ \textbf{0.076} \\

Gripper agr. $\uparrow$
& 0.758 $\rightarrow$ \textbf{0.793}
& 0.820 $\rightarrow$ \textbf{0.957} \\
\bottomrule
\end{tabular}
\vspace{-6pt}
\end{wraptable}

Table~\ref{tab:bottleneck_retrieval} shows that PRISM induces substantially stronger behavioral structure. In the cross-backbone comparison, Wan--DINO neighborhood agreement increases from 0.17 to 0.51 on the main view, while the retrieved transitions also become more similar in action and physical motion. The wrist view shows the same trend, with neighborhood agreement increasing from 0.11 to 0.33 and action RMSE decreasing from 0.37 to 0.27. In a separate DINOv3-only objective ablation (Appendix~\ref{app:analysis_objective}, Table~\ref{tab:appendix_objective_neighbors}), full PRISM reduces action RMSE relative to reconstruction-only training from 0.414 to 0.356 on the main view and from 0.326 to 0.266 on the wrist view, with corresponding improvements in end-effector and gripper similarity. Thus, inverse-dynamics supervision contributes behavioral structure beyond compact feature reconstruction alone.

Taken together, Tables~\ref{tab:bottleneck_cka} and \ref{tab:bottleneck_retrieval} show that PRISM reorganizes native visual features at both global and local scales. Across heterogeneous visual backbones, PRISM increases agreement in transition-specific geometry and produces neighborhoods that are more coherent in terms of actions and physical motion. The matched reconstruction-only comparison further shows that inverse-dynamics supervision contributes directly to this behavioral organization, rather than the effect arising from compact reconstruction alone.

\section{Discussion} 
\label{sec:discussion} 
We presented MiniWAM, which replaces dense native future prediction with compact PRISM targets learned from current--future transitions through inverse dynamics and reconstruction. Across VAE and DINOv3 features, this improves policy performance while substantially reducing training cost, highlighting the potential of learning compact predictive targets for more efficient world--action modeling. \textbf{Limitations and future work.} PRISM is currently learned separately for each downstream setting. Future work could pretrain a shared PRISM encoder on larger and more diverse robot datasets to study transfer across tasks, environments, and embodiments. In addition, our two-stage formulation freezes PRISM before world--action training; future work could explore jointly or partially adapting the encoder in Stage~2 to further improve performance.

\subsection*{AI use statement}

We used generative AI tools to assist with refinement of experimental and analysis protocols, code implementation and debugging, interpretation of experimental results, and polishing of manuscript text. The authors directed the research and retain responsibility for the methodology, experimental validation, and all claims and content presented in this paper.

\subsection*{Ethics statement}

This work studies efficient robot policy learning using existing datasets and simulation benchmarks. Performance on these benchmarks does not establish safety or reliability in real-world environments, where distribution shifts and policy errors may cause physical harm. Deployment on physical robots requires task-specific safety validation, appropriate operational constraints, and human oversight. Use and redistribution of the datasets and pretrained models remain subject to their respective licenses and usage terms. Our planned code release is intended to support transparent evaluation and further research into reliable robot learning.

\subsection*{Reproducibility statement}

The main paper and appendix describe the method and experimental protocols.
Section~\ref{sec:method} specifies the two-stage learning procedure, while
Appendices~\ref{app:architecture}--\ref{app:ablations} provide architecture
details, training hyperparameters, dataset and evaluation settings, and
controlled-comparison protocols. Appendix~\ref{app:representation_analysis}
details the deterministic sampling, representation readouts, and analysis
procedures. We will publicly release the implementation, experiment
configurations, trained model checkpoints, and evaluation scripts
upon acceptance to facilitate reproduction of our results.

\bibliography{iclr2027_conference}
\bibliographystyle{iclr2027_conference}

\appendix
\input{appendix/architecture}
\input{appendix/training}
\input{appendix/evaluation}
\input{appendix/ablations}
\input{appendix/representation_analysis}

\end{document}

%% file: math_commands.tex
\usepackage{amsmath,amsfonts,bm}

\def\eqref#1{equation~\ref{#1}}

\def\1{\bm{1}}

\DeclareMathAlphabet{\mathsfit}{\encodingdefault}{\sfdefault}{m}{sl}
\SetMathAlphabet{\mathsfit}{bold}{\encodingdefault}{\sfdefault}{bx}{n}



%% file: appendix/architecture.tex
\section{MiniWAM and PRISM Architecture}
\label{app:architecture}

\paragraph{Visual features and bottleneck.}
For each camera, the frozen Wan2.1 VAE receives RGB resized bilinearly to $224\times224$ and scaled to $[-1,1]$. A causal encoding of the current frame and 16 future frames yields one current and four future $16\times28\times28$ latent blocks. The alternative frozen DINOv3 ViT-B/16 (LVD-1689M, revision \texttt{5931719e67bbdb9737e363e781fb0c67687896bc}) uses the pinned Hugging Face image processor at $224\times224$: RGB input is converted to $[0,1]$, then the processor resizes, crops and normalizes according to its saved processor settings with rescaling disabled. We take final-layer-normalized patch tokens, excluding class and register tokens and without L2 normalization: $17\times196\times768$ for current plus 16 future frames. DINO computation is FP32 and features cross the policy boundary in BF16. Its current map is projected $768\to64$ by a $1\times1\times1$ convolution; the 16 future maps are projected likewise and temporally compressed by two $(3,1,1)$ convolutions of stride $(2,1,1)$, giving four future maps. No future frame is selected by subsampling.

The VAE transition grid $(5,28,28)$ passes through three 3D convolution--GroupNorm--SiLU blocks with output channels $(32,64,128)$, kernels $(1,3,3),(1,3,3),(3,3,3)$, strides $(1,2,2),(1,2,2),(2,2,2)$, and same padding. It produces $(T_L,H_L,W_L)=(3,4,4)$. DINO instead uses output channels $(64,128,128)$ and strides $(1,2,2),(1,2,2),(2,1,1)$ after its adapter and temporal compressor. This produces $(3,4,4)$ from $(5,14,14)$. A $1\times1\times1$ head sets $D_L=64$. Thus $T_L\times H_L\times W_L\times D_L$ denotes temporal positions, spatial rows, spatial columns, and channels \emph{per camera}; the default is $3\times4\times4\times64$, or 48 world tokens per camera. Camera-role encoders share weights within a role; camera grids are concatenated as tokens and assigned separate tiled rotary-position bands.

\paragraph{Feature-grid compression versus transformer sequence length.}
\label{app:compression}
The approximately $65\times$ compression refers to the number of native future feature-grid locations replaced by PRISM targets, before the world transformer's additional patch embedding. At the default 16-frame future horizon and $224\times224$ image resolution, counts per camera are:
\begin{center}
\begin{tabular}{lrrr}
\toprule
Quantity & Native Wan-VAE & Native DINOv3 & PRISM \\
\midrule
Target grid & $4\times28\times28$ & $16\times14\times14$ & $3\times4\times4$ \\
Feature-grid locations & 3,136 & 3,136 & 48 \\
Channels per location & 16 & 768 & 64 \\
Target scalars & 50,176 & 2,408,448 & 3,072 \\
Future/latent transformer tokens & 784 & 784 & 48 \\
\bottomrule
\end{tabular}
\end{center}
Thus PRISM reduces target-grid locations by $3136/48\approx65.3\times$ for either backbone. The native-prediction baselines additionally patch their feature maps with stride $(1,2,2)$ before the world transformer, whereas PRISM uses one token per latent location. Consequently, the reduction in future/latent \emph{transformer tokens} is $784/48\approx16.3\times$, not $65\times$. Scalar-target compression is $16.3\times$ for Wan-VAE and $784\times$ for DINOv3 because channel counts differ. These counts exclude current-observation conditioning, action tokens, and text context; they do not imply the same reduction in total sequence length or wall-clock training cost. PRISM targets encode a current--future transition, whereas the native prediction targets contain the future features. The ratios here describe the default bottleneck and horizon, not every capacity or horizon ablation.

\paragraph{Posterior and reconstruction.}
Separate $1\times1\times1$ heads output $\mu$ and $\log\sigma^2$ for a diagonal Gaussian. Stage~1 uses reparameterized samples for inverse action prediction and, when enabled, reconstruction; Stage~2 freezes the trained encoder and takes $\mu$ as its deterministic target. The VAE feature decoder trilinearly upsamples the bottleneck to the future target grid, then applies $3\times3\times3$ convolutions $64\to64\to16$ with GroupNorm and SiLU between them. The DINO decoder has the same structure but outputs 768 channels at the native future patch-grid and frame resolution. Both reconstruct features, rather than RGB, and are discarded for policy inference.

\paragraph{Mixture of Transformers.}
The 0.25B model uses 12 MoT layers. Its world and action DiT experts each have width 768, FFN width 3072, 12 heads of dimension 64, and their own input/output projections, feed-forward blocks and timestep modulation. At each layer their queries, keys and values are concatenated for mixed attention, then split for expert-specific residual and feed-forward computation. Parameter totals count both experts, context projections, and active heads; the frozen visual backbone and text encoder are external feature extractors. The implementation reports approximately 247.5M policy parameters for this configuration. Native-feature arms replace the world-token projection/head with a video expert matched to the native target, so parameter counts should be compared using the run manifests rather than assumed identical.

\paragraph{Attention masks.}
Rows are queries and columns are keys; $1$ means visible. Within each block attention is unrestricted. $z_0$ denotes clean current-observation tokens (absent from the Stage-1 inverse-dynamics MoT).
\begin{center}
\begin{tabular}{lcccccc}
\toprule
& \multicolumn{3}{c}{Stage 1} & \multicolumn{3}{c}{Stage 2 joint} \\
Query $\backslash$ key & $z_0$ & $L$ & $a$ & $z_0$ & $L$ & $a$ \\
\midrule
$z_0$ & --- & --- & --- & 1 & 0 & 0 \\
$L$ & --- & 1 & 0 & 1 & 1 & 1 \\
$a$ & --- & 1 & 1 & 1 & 1 & 1 \\
\bottomrule
\end{tabular}
\end{center}
In Stage~1, actions attend to the privileged $L$ but $L$ cannot attend to actions: the task is to infer actions from a known transition, not to predict future states. In Stage~2 joint training, $L$ and actions attend bidirectionally. Missing-camera tokens and padded context keys are additionally masked.

%% file: appendix/training.tex
\section{Training and Inference Details}
\label{app:training}

Table~\ref{tab:appendix_hparams} consolidates the common settings. An epoch refers to one pass over the corresponding extracted training windows; the LIBERO schedule is 16,260 and 54,200 optimizer updates for 15 and 50 epochs, respectively. RoboTwin uses 6 and 25 epochs. The epoch counts are the reported-run settings; the saved checkpoint configuration is the final provenance for each run.
\begin{table}[h]
\centering\small
\begin{tabular}{lll}
\toprule
Setting & Stage 1 & Stage 2 \\
\midrule
Optimizer; peak LR; decay & AdamW; $5\cdot10^{-4}$; 0 & same \\
LR schedule; warmup & cosine; 5\% of updates & same \\
Global batch; precision; seed & 256; BF16; 42 & same \\
LIBERO duration & 15 epochs & 50 epochs \\
RoboTwin duration & 6 epochs & 25 epochs \\
LIBERO action / visual horizon & 16 / 16 & same \\
RoboTwin action / visual horizon & 24 / 24 & same \\
$\lambda_a,\lambda_L$ & $1,0$ & $1,1$ \\
$\beta$; KL warmup & $10^{-6}$; 1,000 updates & $0$; none \\
$\lambda_{\mathrm{rec}}$ & $1$ & $0$ \\
Train/infer flow shift; steps & $5/5$; 1,000 train bins & same \\
Inference & --- & 20 denoising steps \\
Replan interval (environment steps) & --- & LIBERO 10; RoboTwin 24 \\
\bottomrule
\end{tabular}
\caption{Shared 0.25B training settings. Microbatch and gradient accumulation vary with GPU count while their product with GPU count remains 256.}
\label{tab:appendix_hparams}
\end{table}

Both branches use independent Gaussian noise and, in joint training, independent sampled timesteps. For $u\sim\mathcal U(0,1)$, the shifted flow schedule uses $\sigma=5u/(1+4u)$ and timestep $1000\sigma$. The input is $\widetilde y=(1-\sigma)y+\sigma\epsilon$ and the velocity target is $\epsilon-y$. The loss weight is a normalized, zero-minimum version of $\exp[-2(\sigma-1/2)^2]$. Inference traverses the same shifted schedule from $u=1$ to $0$ with Euler steps. In Stage~1 the world tokens remain clean and only action flow loss is active. KL weight rises linearly from zero to $\beta$ over 1,000 updates. Stage~2 jointly denoises action and world tokens, using the frozen Stage-1 encoder's posterior mean as the target. It initializes a new policy from the configured seed; the Stage-1 inverse-dynamics policy and feature decoder are not transferred.

The PRISM runs use $\lambda_{\mathrm{rec}}=1$ for both Wan-VAE and DINOv3. Reconstruction-only controls use weight $1$. The Stage-2 decoder is inactive with $\lambda_{\mathrm{rec}}=0$; where retained in a checkpoint for compatibility it is frozen. The trained encoder alone supplies Stage-2 targets.

For LIBERO, a 16-action chunk is predicted every 10 environment steps. The RoboTwin policies reported in Table~\ref{tab:sota} use 24-step action and visual prediction horizons and execute the complete 24-action chunk before replanning. The separate efficiency benchmarks use visual horizons of 16 and 32 with a fixed 16-step action horizon to compare matched workloads. Initial action/world noise is sampled once per policy call and integrated for 20 inference steps. Current visual features and the instruction embedding are recomputed or retrieved for each replanning call. The frozen instruction cache uses the exact UMT5 prompt and tokenization path as online encoding; packed active BF16 rows are checked with \texttt{torch.equal} against the scalar encoder output, and vanilla LIBERO cache matches are checked exactly. The cache is described in Appendix~\ref{app:evaluation}. Training uses BF16 and the pinned Conda environment: Python 3.10, PyTorch 2.11.0, Transformers 4.57.2, Accelerate 1.14.0, NumPy 1.26.4, and Einops 0.4.1. The LIBERO simulator dependencies include MuJoCo 3.3.2 and Robosuite 1.4.0.

%% file: appendix/evaluation.tex
\section{Datasets and Evaluation Protocols}
\label{app:evaluation}

\subsection{Standard LIBERO}
We evaluate all four suites (\texttt{libero\_10} (\textsc{LIBERO-Long}),
\texttt{libero\_spatial},
\texttt{libero\_object},
\texttt{libero\_goal}), ten tasks each and 50 episodes per task: 2,000 episodes total. Trial $i$ takes the $i$th saved benchmark initial state. The base seed is 42; Python, NumPy, PyTorch, and environment trial seeds use $42+i$. The policy call receives diffusion seed $42+i$ at every replan in that trial, so the seed is reused across replans. The maximum active horizon is 700 steps for \texttt{libero\_10} and 400 for the other suites, plus 30 initial no-op settling steps. Success is the environment's task-success predicate. RGB observations follow the selected visual encoder's preprocessing in Appendix~\ref{app:architecture}; the policy predicts normalized 16-step action chunks, denormalizes using the training-set action normalization statistics, and executes the first ten actions before replanning. The normalized action prediction is clipped to its training support before denormalization; the gripper channel is mapped to the simulator sign convention.

\subsection{LIBERO-Plus} 
The official LIBERO-Plus fork is pinned to commit \texttt{\seqsplit{4976dc30028e805ff8094b55501d532c48fec182}}; assets use revision \texttt{\seqsplit{dd2bd61b7d9a6fef1abc52d606e983b41886a149}}, with archive SHA-256 \texttt{\seqsplit{96764a4bfbdaea98d4411598caeab235458318fe0f549611b93d1a323027b3cf}}. The full robustness evaluation contains 10,030 perturbed task IDs, one episode each. We additionally construct a fixed 2,000-task ablation subset using seed 42. The subset covers every nonempty suite $\times$ perturbation category $\times$ difficulty stratum (141 strata), allocates the remaining slots proportionally, and ranks candidates by stable SHA-256. Missing difficulty is retained as \emph{unclassified}. We report the full population as \textsc{LIBERO-Plus} and the fixed 2,000-task subset as \textsc{LIBERO-Plus-2K}.

The text cache retains the raw fork language, a clean model-facing instruction, and the matched vanilla LIBERO instruction for auditing. For non-language perturbations, the clean string is the vanilla instruction, excluding filename-derived camera, state, texture, and layout suffixes. For the language category, the clean string preserves the deliberate paraphrase. Only the clean string is encoded. The cache checks exact BF16 tensor equivalence to online scalar UMT5 encoding and exact context/mask equivalence for mapped vanilla instructions; task IDs, text fields, revisions and hashes remain in the evaluation record. The official task-ID manifest and pinned assets determine the initial states and perturbations.

\subsection{RoboTwin 2.0}
The training split is the official XPolicyLab-format \texttt{demo\_clean} data: 50 tasks, 50 demonstrations per task, episode IDs \texttt{0000000}--\texttt{0000049}. It uses head, left-wrist, and right-wrist RGB cameras; the two wrist-camera streams share encoder weights. Actions are 14-dimensional absolute joint positions, packed as six joints and a gripper for each arm. The official multitask evaluator is run separately under \texttt{demo\_clean} and \texttt{demo\_randomized}, with seed 0 and the benchmark's online valid-case/expert filter. The official leaderboard protocol evaluates 100 valid episodes per task under each of \texttt{demo\_clean} and \texttt{demo\_randomized}. We use the benchmark's online expert-feasibility filter to select valid evaluation cases, and report the mean task success for each condition together with their equal-weight average. Smaller frozen-manifest runs (e.g., 20 cases per task per condition, 2,000 total) are diagnostic subsets and are not substituted for the full leaderboard protocol. Frozen replay verifies recorded seeds, scenes and instructions; official mode performs online selection. The RoboTwin MiniWAM policies in Table~\ref{tab:sota} are trained with a 24-frame visual prediction horizon and a 24-step action horizon. Evaluation executes all 24 predicted actions before replanning, using the checkpoint's action and proprioception statistics. The horizon-16/32 efficiency measurements are separate profiling workloads.

\subsection{Provenance of Externally Reported Baselines}
\label{app:sota_provenance}

Table~\ref{tab:sota_provenance} gives the per-cell sources for all entries
marked with $\dagger$ in Table~\ref{tab:sota}. Unmarked entries in the main
table are our evaluations of released checkpoints. A dash means that the
corresponding value is not used in the main comparison.

\begin{table}[h]
\centering
\caption{\textbf{Sources of externally reported results in Table~\ref{tab:sota}.}
Values are success rates (\%).}
\label{tab:sota_provenance}
\small
\setlength{\tabcolsep}{4pt}
\begin{tabular}{@{}lp{0.21\textwidth}p{0.21\textwidth}p{0.21\textwidth}@{}}
\toprule
Method & \textsc{LIBERO} & \textsc{LIBERO-Plus} & \textsc{RoboTwin 2.0} \\
\midrule
$\pi_0$-FAST
& 85.5~\citep{Fei_2026_CVPR}
& 61.6~\citep{Fei_2026_CVPR}
& -- \\
UniVLA
& 95.2~\citep{UniVLA}
& 42.9~\citep{Fei_2026_CVPR}
& -- \\
StarVLA-OFT
& 96.6~\citep{community2026starvla}
& --
& 24.8~\citep{robotwin_leaderboard} \\
Xiaomi-Robotics-0
& 98.7~\citep{cai2026xiaomi}
& --
& 40.5~\citep{robotwin_leaderboard} \\
WorldVLA
& 79.1~\citep{cen2025worldvla,Fei_2026_CVPR}
& 25.0~\citep{Fei_2026_CVPR}
& -- \\
AHA-WAM
& --
& --
& 33.8~\citep{robotwin_leaderboard} \\
\bottomrule
\end{tabular}
\end{table}

For UniVLA, we use the 42.9 reported in the archival CVPR 2026
LIBERO-Plus paper. WorldVLA reports 81.8 on standard \textsc{LIBERO} with its
$512\times512$ model, but the LIBERO-Plus study
evaluates the $256\times256$ checkpoint, which obtains 79.1 on standard
\textsc{LIBERO} and 25.0 on \textsc{LIBERO-Plus}; we report this matched pair.
The externally reported \textsc{RoboTwin} values correspond to the official clean2clean/clean2random leaderboard track, where policies are trained only on clean demonstrations. This differs from results reported by some methods under clean+randomized training. For example, AHA-WAM reports 92.8 after training on both clean and randomized data \citep{cai2026aha}, whereas the clean-trained leaderboard entry used in Table~\ref{tab:sota} is 33.8.

%% file: appendix/ablations.tex
\section{Controlled Baselines and Ablations}
\label{app:ablations}

Within each visual-feature family, all LIBERO arms use the same four-suite data, current observations, 16-step action/visual horizons, action expert width, training seed, global batch, 54,200 Stage-2 optimizer updates, and evaluation episodes. The world branch changes with its prediction target; its input/output projections and token count therefore change, and exact parameter counts are reported from each run. \emph{Without co-training} sets the direct world-flow coefficient to zero but retains the sampled world tokens and joint attention, so the action loss can still update world-expert parameters. \emph{Native future-feature co-training} predicts all four future Wan-VAE blocks or all 16 DINOv3 patch grids per camera, with no PRISM encoder. \emph{PRISM co-training} predicts the frozen transition encoder's posterior-mean grid. All three share the same action objective; native and PRISM arms set world-loss weight to one.

For the Stage-1 objective ablation, reconstruction-only trains the encoder and feature decoder with feature MSE and KL but no inverse action loss; inverse-dynamics-only trains the encoder with action flow and KL but no decoder loss; inverse dynamics plus reconstruction enables both. The resulting encoders are frozen and each supplies the same Stage-2 architecture and schedule. A reconstruction-only \emph{encoder} arm, if distinguished in a table, keeps the policy at its seeded initialization during Stage~1. Capacity variants change only the encoder bottleneck grid or channel width, with matched data and optimizer budgets. The table shorthand $4\times4$, $2\times2$, and $1\times1$ specifies $H_L\times W_L$; the full bottleneck is $T_L\times H_L\times W_L$ with $T_L=3$ at the 16-frame horizon.

\paragraph{Efficiency measurement.}
\label{app:efficiency_accounting}
The \texttt{training\_core} boundary starts from resident tensors at the
extracted-feature boundary and measures complete forward, backward, and
optimizer updates, excluding disk I/O, data loading, collation, and
host-to-device transfer. The \texttt{training\_pipeline} boundary additionally
includes these data-pipeline costs. Throughput is measured on one NVIDIA RTX
PRO 6000 Blackwell Max-Q Workstation Edition GPU with 96\,GB memory, BF16,
global batch 256, and a memory-calibrated microbatch. The main-table
\textsc{LIBERO} runs use one warmup and three measured updates, while
\textsc{RoboTwin} runs use two warmup and six measured updates. All timings use
the same pinned software environment as training.

Efficiency experiments fix the action horizon at 16 and evaluate visual
horizons of 16 and 32. Stage-2 speedup is
$S_{\mathrm{S2}}=t_{2,n}/t_{2,p}$ under matched horizons, camera counts, and
hardware. We additionally account for the one-time PRISM Stage-1 cost.
Measured Stage-1 and Stage-2 PRISM update times are nearly identical
(Stage-1/Stage-2 ratios of 0.97 on \textsc{LIBERO} and 1.07 on
\textsc{RoboTwin}), so we use a common per-update PRISM cost for this
accounting. With 15/50 Stage-1/Stage-2 epochs on \textsc{LIBERO} and 6/25 on
\textsc{RoboTwin},
\[
S_{\mathrm{total}}
=
\frac{S_{\mathrm{S2}}}{1+N_1/N_2}
=
\begin{cases}
S_{\mathrm{S2}}/1.30 & \textsc{LIBERO},\\
S_{\mathrm{S2}}/1.24 & \textsc{RoboTwin}.
\end{cases}
\]
All speedups are computed before rounding.

\paragraph{DINOv3 training pipeline throughput.}
\label{app:pipeline_efficiency}
Table~\ref{tab:appendix_pipeline_efficiency} complements the isolated compute results in Table~\ref{tab:training_efficiency} with throughput through the training data pipeline. Both arms use extracted visual features; PRISM additionally loads cached posterior-mean targets from its frozen encoder. Pipeline timing includes data loading, collation, host-to-device transfer, forward and backward passes, and optimizer updates. Offline feature and target extraction are excluded. The LIBERO native pipeline runs use one warmup and three measured updates; cached PRISM and RoboTwin pipeline runs use two warmup and six measured updates. LIBERO uses one main and one wrist view, while RoboTwin uses one main and two wrist views. PRISM retains 96.9--99.2\% of its core throughput, with pipeline Stage-2 speedups ranging from $3.95\times$ to $9.60\times$. Thus the reduction in world-modeling compute also translates into faster updates through the data pipeline.

\begin{table}[ht]
\centering
\small
\setlength{\tabcolsep}{4pt}
\caption{\textbf{DINOv3 training throughput through the data pipeline.} Samples per second on one RTX PRO 6000 GPU at global batch 256. Core values reproduce the compute measurements in Table~\ref{tab:training_efficiency}; pipeline values include data loading and transfer. Speedup compares PRISM and native Stage-2 pipeline throughput.}
\label{tab:appendix_pipeline_efficiency}
\begin{tabular}{@{}lcrrrrr@{}}
\toprule
Benchmark & Visual & \multicolumn{2}{c}{Native} & \multicolumn{2}{c}{PRISM} & Pipeline \\
& horizon & Core & Pipeline & Core & Pipeline & speedup \\
\midrule
LIBERO & 16 & 36.17 & 35.16 & 143.25 & 138.81 & $3.95\times$ \\
& 32 & 15.55 & 15.26 & 126.09 & 123.08 & $8.07\times$ \\
\midrule
RoboTwin 2.0 & 16 & 21.48 & 21.11 & 93.57 & 92.86 & $4.40\times$ \\
& 32 & 8.61 & 8.41 & 81.47 & 80.74 & $9.60\times$ \\
\bottomrule
\end{tabular}
\end{table}

%% file: appendix/representation_analysis.tex
\section{Representation-Analysis Protocols}
\label{app:representation_analysis}

\subsection{Cohort and representation readouts}
\label{app:analysis_cohort}

We use task-indexed \textsc{LIBERO-90} demonstrations, disjoint in task ID from the \textsc{LIBERO-40} data used to train the analyzed transition encoders. A fixed seed-42 manifest contains 4,500 transitions: five examples in every one of the $90\times10$ task-by-progress cells. The nested CKA subset takes two examples per cell, for 1,800 transitions. Each row records an immutable sample ID, official task ID, source demonstration, anchor frame, and normalized progress $t/\max(1,T-1)$, where $T$ is the episode length. The analysis verifies the sample-ID order in every source and posterior bank and uses the same paired rows at every representation stage. Main and wrist cameras are analyzed separately.

Every representation is derived from the same physical current RGB frame and 16 future RGB frames. The native Wan-VAE readout spatially averages each 16-channel latent block and concatenates one current with four causal future blocks, giving $5\times16$ values per transition. The native DINOv3 readout spatially averages patch features independently for the current and 16 future frames, giving $17\times768$ values. For each trained PRISM encoder we flatten the deterministic posterior mean $L_\mu$ in channel--time--space order, retaining the full $64\times3\times4\times4=3{,}072$ values. Thus the physical input window is matched, but native temporal resolution and feature dimension differ. Native features are spatially pooled whereas the primary PRISM readout retains its coarse spatial grid; pooled $L_\mu$ and intermediate features are sensitivity analyses.

The cross-backbone table uses the matched Wan and DINO Stage-1 inverse-dynamics-plus-reconstruction checkpoints, both with reconstruction weight $\lambda_{\mathrm{rec}}=1$, consistent with the policy-training configuration in Appendix~\ref{app:training}. The analysis does not mix inverse-dynamics-only posterior banks with the matched reconstruction-enabled pair.

\subsection{Cross-backbone CKA}
\label{app:analysis_cka}

For exactly paired activation matrices $X\in\mathbb R^{n\times d_X}$ and $Y\in\mathbb R^{n\times d_Y}$, conventional linear centered kernel alignment (CKA) is
\begin{equation}
\operatorname{CKA}_{\mathrm{lin}}(X,Y)=
\frac{\|X_c^\top Y_c\|_F^2}
{\sqrt{\|X_c^\top X_c\|_F^2\|Y_c^\top Y_c\|_F^2}},
\label{eq:appendix_linear_cka}
\end{equation}
where columns are centered across samples. The table reports the debiased linear estimator. With linear Gram matrices $K=XX^\top$ and $G=YY^\top$, first set their diagonals to zero and compute the U-statistic
\begin{equation}
\widehat{\mathrm{HSIC}}_U(K,G)=\frac{1}{n(n-3)}\left[
\operatorname{tr}(KG)+
\frac{(\mathbf1^\top K\mathbf1)(\mathbf1^\top G\mathbf1)}{(n-1)(n-2)}-
\frac{2\mathbf1^\top KG\mathbf1}{n-2}\right].
\label{eq:appendix_hsic_u}
\end{equation}
We divide this by $\sqrt{\widehat{\mathrm{HSIC}}_U(K,K)\widehat{\mathrm{HSIC}}_U(G,G)}$. This reduces high-dimensional estimator bias: the full-grid representation has more dimensions than paired samples. Exact feature-space sufficient statistics or exact sample-Gram terms are used according to the smaller intermediate; separately normalized minibatch CKA values are never averaged. We do not standardize individual feature dimensions, because anisotropic rescaling changes the geometry. Conventional CKA is retained as a sensitivity.

We compute CKA under three distinct conditioning protocols:
\begin{itemize}
\item \textbf{Global.} Use all paired rows without task-specific preprocessing. Ordinary sample centering is implicit in conventional CKA; the debiased U-statistic is computed on the original rows. This retains task, scene, progress, current-observation, and transition variation.
\item \textbf{Within-task.} For each official task $g$, separately center each backbone and stage: $X'_i=X_i-|\mathcal I_g|^{-1}\sum_{j\in\mathcal I_g}X_j$ for $i\in\mathcal I_g$, and analogously for $Y$. Compute debiased CKA over the combined residual rows. This removes task-wise mean offsets, not within-task scene variation.
\item \textbf{Task/current deconfounded.} For each backbone and representation stage, predict its feature vector from official-task one-hot, normalized progress, squared progress, and that backbone's spatially pooled current-only native feature. Compute debiased CKA between the resulting held-out residual matrices. This removes only structure linearly predictable by the specified nuisance model; it does not guarantee complete removal of current-scene information.
\end{itemize}

For deconfounding, each target matrix has a separate multi-output ridge regression. Predictors and targets are standardized using training-fold statistics only. Complete demonstrations, identified by dataset and episode ID, are assigned deterministically to five approximately balanced outer folds by hashing the episode key with seed 42. Within each outer training split, three episode-grouped inner folds select one ridge coefficient from $\{10^{-4},10^{-2},1,10^2,10^4\}$ by mean validation MSE. A transition's nuisance prediction is therefore obtained from a model that has seen no transition from its demonstration. The same fixed paired-row cohort and fold protocol are used before and after PRISM. Backbone-specific fits are necessary because the Wan and DINO current-only features have different dimensions and scales.

The before/after comparison contrasts Wan--DINO CKA on native transition readouts with CKA on the two trained full-grid posterior means. We also evaluate seed-42 random Wan and DINO transition encoders with the same backbone-specific adapters, encoder architecture, and output grid. The random-pair CKA is computed under \emph{all three} conditioning protocols; a trained value that does not exceed this control cannot be attributed to learned alignment. Only one trained checkpoint pair and one random initialization are analyzed, so the results are checkpoint-specific rather than estimates of training-seed variation.

The primary global before/after difference has a 1,000-replicate fixed-task hierarchical bootstrap interval: all 90 tasks remain in each replicate, demonstrations are sampled with replacement within task, and selected rows from a demonstration move together. A 100-permutation null permutes DINO rows within task, applying the same permutation at the native and PRISM stages. Within-task, deconfounded, and trained-minus-random comparisons are reported as paired \emph{point estimates}: a naive clustered percentile interval for nonlinear debiased CKA was poorly centered, while a full deconfounded interval would additionally require refitting the nested nuisance models inside each resample. These analyses neither establish causal mediation nor imply coordinate-wise equivalence.

\subsection{Cross-backbone neighborhood retrieval}
\label{app:analysis_neighbors}

The 4,500-row bank is split into 882 fixed queries and 3,618 gallery transitions with no shared source demonstration between query and gallery. For each query, cosine-distance retrieval excludes every gallery transition with the same official task ID. We retrieve the top five from identical query and gallery IDs using Wan native, DINO native, Wan-PRISM full-grid, and DINO-PRISM full-grid readouts. Jaccard@5 is the per-query intersection-over-union of the Wan- and DINO-retrieved ID sets, averaged over queries. Full-grid PRISM is primary; spatially pooled PRISM is a sensitivity.

Behavioral metrics compare each query with each retrieved transition. Action RMSE uses normalized 16-step action chunks; end-effector error is the Euclidean distance between current-to-future Cartesian displacement vectors in metres; gripper agreement is the fraction of action timesteps with matching command signs. We compute each metric separately for Wan- and DINO-retrieved neighbors, then report their symmetric mean. The DINO reconstruction-only and inverse-dynamics-plus-reconstruction comparison uses identical transition IDs, query/gallery split, normalization, and retrieval rules.

\subsection{DINOv3 objective ablation}
\label{app:analysis_objective}

Table~\ref{tab:appendix_objective_neighbors} isolates the contribution of inverse-dynamics supervision using DINOv3 bottlenecks with the same latent width (64) and posterior grid ($3\times4\times4$). Both objectives reconstruct native features; full PRISM additionally uses inverse dynamics, with reconstruction weight 1. We use the same 882 queries, 3,618 gallery transitions, and leave-task-out cosine top-5 protocol described above. These are DINO-only metrics, unlike Table~\ref{tab:bottleneck_retrieval}, which compares raw transitions against PRISM and averages the Wan and DINO retrieval metrics.

\begin{table}[t]
\centering
\caption{\textbf{Effect of inverse-dynamics supervision on neighborhood behavior.} Matched DINOv3 bottlenecks on unseen \textsc{LIBERO-90} tasks; both retain the full posterior-mean grid.}
\label{tab:appendix_objective_neighbors}
\small
\begin{tabular}{llccc}
\toprule
View & Objective & Action RMSE $\downarrow$ & EEF Err. (m) $\downarrow$ & Grip. Agr. $\uparrow$ \\
\midrule
Main & Reconstruction only & 0.414 & 0.090 & 0.747 \\
& PRISM (I.D.\ + recon.) & \textbf{0.356} & \textbf{0.068} & \textbf{0.802} \\
\midrule
Wrist & Reconstruction only & 0.326 & 0.087 & 0.907 \\
& PRISM (I.D.\ + recon.) & \textbf{0.266} & \textbf{0.078} & \textbf{0.975} \\
\bottomrule
\end{tabular}
\end{table}